\documentclass{article}

\usepackage{natbib}

\usepackage[utf8]{inputenc} 
\usepackage[T1]{fontenc}    
\usepackage{hyperref}       
\usepackage{url}            
\usepackage{booktabs}       
\usepackage{amsfonts}       
\usepackage{nicefrac}       
\usepackage{microtype}      
\usepackage{xcolor}         
\usepackage{mwe}
\usepackage{algorithm}
\usepackage{algorithmic}

\usepackage{tikz}

\usepackage{url}
\usepackage{multirow}
\usepackage{adjustbox}
\usepackage{footnote}
\usepackage{booktabs}
\usepackage{graphicx}
\usepackage[none]{hyphenat}
\usepackage{makecell}
\usepackage{framed}
\usepackage{tabto}
\usepackage{hyperref}
\usepackage{tipa}
\usepackage{subcaption}

\usepackage{amsmath}

\usepackage{ulem}
\usepackage{makecell}

\def\Y{\mathbf{Y}}
\def\W{\mathbf{W}}

\def\X{\mathbf{X}}
\def\x{\mathbf{x}}
\def\Y{\mathbf{Y}}
\def\y{\mathbf{y}}
\def\real{\rm I\!R}

\def\U{\mathbf{U}}
\def\u{\mathbf{u}}
\def\V{\mathbf{V}}
\def\v{\mathbf{v}}
\def\SSigma{\boldsymbol{\Sigma}}
\DeclareMathOperator*{\argmax}{argmax} 

\title{Accelerating the Low-Rank Decomposed Models}

\author{
    Habib Hajimolahoseini\thanks{Ascend Team, Toronto Research Center, Huawei Technologies, Toronto, Canada} 
    \and 
    Walid Ahmed\footnotemark[1]
    \and 
    Austin Wen\footnotemark[1]
    \and 
    Yang Liu\footnotemark[1]
}
\begin{document}

\maketitle

\begin{abstract}
    Tensor decomposition is a mathematically supported technique for data compression. It consists of applying some kind of a Low Rank Decomposition technique on the tensors or matrices in order to reduce the redundancy of the data. 
    However, it is not a popular technique for compressing the AI models duo to the high number of new layers added to the architecture after decomposition. Although the number of parameters could shrink significantly, it could result in the model be more than twice deeper which could add some latency to the training or inference. In this paper, we present a comprehensive study about how to modify low rank decomposition technique in AI models so that we could benefit from both high accuracy and low memory consumption as well as speeding up the training and inference. 

\end{abstract}

\section{Introduction}
\label{sec:intro}
With the fast evolution of AI processors used for training the Deep Learning models, the community's focus is moving towards larger and larger models with millions of trainable parameters \citep{ahmed2023speeding, hajimolahoseini2024single, hajimolahoseini2024methods, hajimolahoseini2023methods, hajimolahoseini2021compressing}. 
Tuning all of these parameters consumes a huge portion of memory with a huge and exponentially growing computational complexity during both training and inference \citep{dean2012large}.
For example, ResNet-152, which is one of the most widely used Convolutional Neural Network (CNN), has more than 60 million parameters and over 11 billion FLOPs. \cite{he2016deep}.
In real-time applications especially when these models are deployed on the smartphones and other embedding devices, memory consumption and computational complexity can raise lots of issues including memory and battery life.
However, studies show that such large AI models may include a lot of redundancy in the data they are keeping in terms of their weight matrices/tensors inside their layer \cite{}. 

\subsection{Related Work}
The computational complexity of CNNs and their memory consumption is dominated by the convolutional and fully connected layers, respectively \cite{cheng2018recent}. 
Different techniques and algorithms are proposed in the literature for compressing and/or accelerating the AI models including: Low Rank Decomposition (LRD), pruning, quantization and knowledge distillation \citep{cheng2017survey, ataiefard2024skipvit, javadi2023gqkva, hajimolahoseini2012extended, hajimolahoseini2008improvement, hajimolahoseini2023training}. 
In deep learning models, the objective of compression is to optimize the computational complexity and memory consumption without changing the network architecture \citep{hajimolahoseini2023swiftlearn}.
However, in knowledge distillation, the architecture of the compressed model may be different from that of the original network.

\subsubsection{Pruning}
The main idea behind pruning is that in deep learning models, there are many parameters that doesn't contribute much regarding the model performance \cite{luo2018thinet}
The goal is to increase the sparsity of such models by removing those unimportant parameters. 
This can help in both reducing the memory usage as well as computational complexity. 
Pruning can be done in different levels including: weight level, vector level, kernel level, group level and filter level \cite{zhang2018systematic, zhuang2018discrimination, mao2017exploring}. 
These methods generally consist of three steps. 
At the first step, the original model is first trained. 
Then, based on a ranking criterion, the weights or filters are pruned.
The remaining model is finally fine-tuned in order to recover the accuracy \cite{li2016pruning}. 
The existing pruning methods are different based on the ranking criteria they offer, which is usually designed manually. 
It is hard most of the times to verify these assumptions about the criteria used for ranking.
For example, a typical assumption used in pruning is that the weights with small norms do not contribute much to the performance.
However, there is no mathematical proof for this assumption. 
Another challenge is that it may take many iterations of pruning and fine tuning until we reach a good ranking criterion \cite{li2019compressing}. 
Therefore, although model pruning is shown to be effective for reducing the memory consumption, it is not always helpful training or inference acceleration \citep{cheng2017survey}


\subsubsection{Quantization}
Network quantization can be used for both compression and acceleration. 
The current methods in this area quantize the models using either scalar, vector or fixed-point quantization techniques \cite{cheng2017quantized}. 
In scalar or vector quantization, the data is represented using a codebook of quantization cores as well as a set of quantization codes for assigning to them \cite{cheng2017survey}. 
On the other hand, fixed-point quantization techniques are focused on quantizing the weights or activations of the model \cite{}. 
Tn these methods, model weights are approximated using a smaller number of bits e.g. 16-bits, 8-bits or even a single bit (binary networks) \citep{wu2016quantized, han2015deep, courbariaux2015binaryconnect, prato2019fully, bie2019simplified}.

\subsubsection{Knowledge Distillation}
Knowledge Distillation (KD) uses a teacher-student framework in order to transfer the knowledge from a larger network (teacher) into a compact and  efficient one (student) by adding and auxiliary loss to imitate softmax outputs or logits from the teacher as a representation of class distributions \citep{hinton2015distilling, rashid2021mate}.
In this method, the architecture of the student could be totally different from that of the teacher. 
The student model tries to mimic the behaviour of the teacher through distilling knowledge from the teachers output or intermediate layers \citep{mirzadeh2020improved}.

The theory behind how KD works is still an open question in the literature, however, there are some works explaining the contribution of adding class similarity information in the output of the teacher (which is reffered to as the \textit{dark knowledge} as well), or regularization effects of the KD loss as potential reasons. 
However, these methods have poor mathematical support and could face some serious limitations in high compression ratios.

 

\subsubsection{Tensor Decomposition}
In contrast with most of the model compression/acceleration techniques, LRD has the well established theoretical foundation with a long history in mathematics \cite{jaderberg2014speeding}. 
The weights of fully connected (FC) layers of deep learning models are 2D  matrices while the filters of convolutional layers are 4D tensors.
Therefore, an appropriate matrix or tensor decomposition technique can be applied to decompose them into smaller ones. 
The goal of Low Rank Decomposition (LRD) is to replace the original tensor/matrix with an approximate tensor/matrix that is close enough to it but with more efficiency in calculations \cite{hajimolahoseinicompressing}.

Singular Value Decomposition (SVD) is the most popular method for decomposing the 2D matrices into 2 smaller ones \cite{van1987matrix}. 
In this approach, each FC layer is replaced by 2 consecutive FC layers  whose weights are calculated from the original matrix using SVD algorithm. On the other hand, for convolutional layers, a higher order version of SVD e.g. Tucker is applied in order to decompose them into multiple components \cite{de2000multilinear, de2000best}.  
LRD will be explained in more details in the next sections. 

Although LRD has lots of benefits, there are some shortcomings that prevent it from being a popular method in deep learning community.
LRD is mostly considered as a type of model compression technique which does not help in terms of acceleration.
This is because it will add more layers to the model architecture which makes the model deeper and deeper.
This could cause latency during both training and inference.
In this paper, we present a comprehensive study about how we can improve the LRD methods in order to use them as a type of model acceleration technique as well. 
What motivates us to work on improving LRD for deep learning models is that it has a lot of benefits which may not be provided by the other techniques including:
\begin{itemize}
    \item It is applied only once during the training and takes only few seconds for decomposing the deep models. Therefore it will not add latency to the total training time. 
    
    \item It has a rich mathematical foundation and is not based on heuristics. Therefore it is a more generic technique which can be applied to different types of models.
    
    \item It does not need heavy pre-training because it can start from a large pre-trained model to initialize the compressed model. This is in contrast to many other techniques in which we should train the compressed model from scratch using random initialization. Therefore when using LRD, a few fine-tuning steps are enough to recover most of the accuracy drop.
    
    \item It has a built-in one-shot knowledge distillation technique because the new weights in the decomposed model are calculated from the original model which transform the knowledge from the original (teacher) to the decomposed (student) model in one single shot
\end{itemize}
In the following sections, the LRD and the our proposed modification strategies are explained.

\section{Accelerating Low Rank Decomposition}
Generally speaking, each convolutional layer in deep learning models include a 4D tensor $\W\in \real^{C\times S\times h\times w}$ of trainable parameters, where $C$ and $S$ represent the number of input and output channels, respectively while $h$ and $w$ are the spatial dimensions of the kernels. 
For $1\times 1$ convolutional and fully connected layer, we can consider $h=w=1$. 
Therefore, the 4D tensor $\W$ could be represented in 2D space as $\W\in \real^{C\times S}$.


In the case of fully connected or $1\times 1$ convolutional layers, the weight matrix $\W$ is decomposed using Singular Value Decomposition (SVD) as follows \cite{hajimolahoseinicompressing}: 
\begin{equation}\label{svd}
\W = \U\SSigma \V^\top=\sum_{i=1}^r{\sigma_i\u_i\v_i^\top},
\end{equation}
where $\U \in \real^{C\times C}$ and $\V \in \real^{S\times S}$ are the orthogonal matrices and $\SSigma \in \real^{C\times S}$ is a diagonal rectangular matrix containing the singular values $\sigma_i>0$ of $\W$ and $r=\min(C,S)$ is called the rank of $\W$ assuming the full-rank.

Using \eqref{svd} doesn't necessarily lead to compression of the layers. 
However, if we only use the first $R<r$ components in \eqref{svd}, the resulting matrix is called a low-rank approximation of $\W$:
\begin{equation}\label{svd2}
\W' = \sum_{i=1}^R{\sigma_i\u_i\v_i^\top}=\U'\SSigma' \V'^\top
\end{equation}
where $\U' \in \real^{C\times R}$ and $\V' \in \real^{S\times R}$ are the new  orthogonal matrices and $\SSigma' \in \real^{R\times R}$ is the new diagonal rectangular matrix.
Based on \eqref{svd2}, $\W'$ can be interpreted as the multiplication of the following two matrices: 
\begin{align}
\W' &= \W_0 \W_1, \quad \W_0 = \U'\sqrt{\SSigma'}, \quad \W_1 = \sqrt{\SSigma'} \V'^\top \label{svd4}
\end{align}
where $\W_0 \in \real^{C\times R}$ and $\W_1 \in \real^{R\times S}$, and $\sqrt{\SSigma}$ is a diagonal of square root of singular values $\sqrt{\sigma_i}.$ 
According to \ref{svd4}, each fully-connected or $1\times 1$ convolutional layer with weight matrix $\W$ could be replaced with two consecutive layers $\W_0$ and $\W_1$, in which the number of parameters could shrink significantly depending on the low rank $R$.

In regular convolutioal layers, the filters are 4D so a higher order version of SVD e.g. Tucker decomposition is needed. 
However, in regular CNNs, since spatial dimensions of kernels ($h$ and $w$) are too small comparing to the feature space dimension (mostly 3 and up to 7 in some models), only the channel related dimensions need to be decomposed.
Hence, for simplicity in notations, assuming that $h=w=k$ we can reshape the tensor $\W\in \real^{C\times S \times h \times w}$ into a 3D tensor as $\W\in \real^{C\times S \times k^2}$.
Now, the Tucker decomposition can be applied as follow:
\begin{equation}\label{tucker2}
\W = \X \times_C \U \times_S \V
\end{equation}
where $\U \in \real^{C\times C}$ and $\V \in \real^{S\times S}$ are unitary matrices and $\X \in \real^{C\times S \times k^2}$ is the core tensor, containing the 1-mode, 2-mode and 3-mode singular values of $\W$. 
Symbols $\times_C$ and $\times_S$ also represent multilinear products between each matrix and the core tensor along dimensions $C$ and $S$, respectively.
Simmilar to \eqref{svd}, Tucker decomposition \eqref{tucker2} can be rewritten in two steps as follows:
\begin{align}\label{tucker2-2}
\Y &= \X \times_C \U = \sum_{i=1}^{r_1}{\u_i\x_i^\top} \\
\W &= \Y \times_S \V = \sum_{i=1}^{r_2}{\y_i\v_i^\top}
\end{align}
in which $\Y\in \real^{C\times r_2\times k^2}$ is the result of multiplying $\U$ with the core tensor $\X$ along dimension $C$.

$r_1=$ and $r_2$

\begin{figure}
    \begin{minipage}{.48\textwidth}
        \centering
        \includegraphics[width=60mm,scale=1]{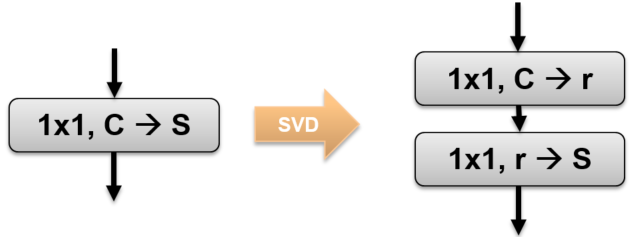}
        \subcaption{SVD}
        \label{fig:1a}
    \end{minipage}
    \begin{minipage}{.48\textwidth}
        \centering
        \includegraphics[width=60mm,scale=1]{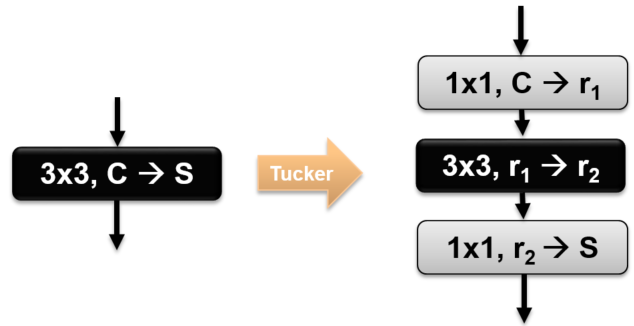}
        \subcaption{Tucker2}
        \label{fig:1b}
    \end{minipage}
    \caption{Low Rank Decomposition of 1x1 and 3x3 convolutional layers. Note that FC layers are treated the same as 1x1 Conv layers. }
    \label{svd_tucker}
\end{figure}

For the regular convolutional layers, since $\W$ has a higher number of dimensions, a higher order version of SVD is applied in order to decompose each layer into 3 or more layers \cite{de2000multilinear, de2000best}. 
In this work, we use Tucker decomposition method which replaces each convolutional layer with weight tensor $\W\in \real^{C\times S\times h\times w}$ into 3 convolutional layers as follows: a 1x1 convolutional layer with weight matrix $\W\in \real^{C\times R_1}$, followed by a regular convolutional layer called the core with weight tensor $\W\in \real^{R_1\times R_2\times h\times w}$, and finally another 1x1 convolutional layer with weight matrix $\W\in \real^{R_2\times S}$ \cite{gusak2019musco}. 
$R_1$ and $R_2$ are the ranks of Tucker decomposition.
The decomposition of 1x1 and 3x3 convolutional layers is illustrated in Fig.\ref{svd_tucker}.

Also, the ranks of decomposition can be calculated using different approaches. 
For more information about different rank selection methods the reader is referred to \cite{gusak2019musco, hajimolahoseinicompressing}. 
Here we calculate the ranks so that each layer has a desired compression ratio.

Most of the papers in the literature use the number of floating point operations per second (FLOPs) as a measure of computational complexity.
However, it can be shown that FLOPs doesn't necessarily reflect the throughput of the network, as models with lower FLOPs could take even more time during training and/or inference.
Table \ref{table0} shows the number of layers, FLOPs, speed in terms of frame per second as well as total number of parameters in ResNet-50, ResNet-101 and ResNet-152 architectures before and after applying LRD to all the layers with 2x compression. 
As seen in this table, the decomposed models have almost 2 times less number of parameters and FLOPs comparing to their original version.
However, according to FPS results during inference, the throughput improvement is $6.8\%$, $10.5\%$, and $13.1\%$ for ResNet-50, ResNet-101 and ResNet-152, respectively which is not that significant comparing to the 2x compression ratio. 
It is because the decomposed models have more than twice number of layers which makes the models deeper. 

\begin{table}[h]
    \centering
    \caption{Statistics of ResNet-50, ResNet-101 and ResNet-152 architectures before and after applying LRD.}
{\footnotesize
    \begin{tabular}{l| c c c c c}
{\bf Model} &{\bf Layers} &{\bf Params (M)} &{\bf FLOPs (B)} &{\bf Train fps} &{\bf Infer fps}\\
\hline
\textbf{ResNet-50} &50 &25.56 &8.23 &346 &1232\\
Vanilla LRD &115 &12.78  &4.67  &367  &1316 \\
\hline
\textbf{ResNet-101} &101 &44.55 &15.68 &207 &713\\
Vanilla LRD &233 &22.21  &8.39  &227  &788 \\

\hline
\textbf{ResNet-152} &152 &60.19 &23.14 &145 &510\\
Vanilla LRD &352 &30.01  &12.11  &162  &577  \\

\end{tabular}
    }
    \label{table0}
\end{table}

\subsection{Appropriate Rank Selection}
The filter dimensions of well-known architectures such as ResNet are selected so that the models could be trained on GPUs in the most efficient way. 
It could be shown that because of the low level design of the calculations on hardware, some specific dimensions such as powers of 2 would result in a more efficient processing on de devices \cite{}. 
That is why all convolutional layers in ResNet models have dimensions that are powers of 2 e.g. 256, 512, etc.
However, this is not necessarily the case after we decompose these models as we calculate the ranks according to a desired compression ratio which may lead to having some odd numbers as the filter dimensions.
This may not be efficient in low-level calculations on hardware.

For example, a convolutional layer with filter dimensions of $[512,512,3,3]$ will be decomposed into 3 convolutional layers of dimensions $[512,309]$, $[309,309,3,3]$ and $[309,512]$ by applying LRD with 2x compression. 
Having tensors with dimensions 309 may not lead to efficient calculations on hardware. 
Therefore, we propose a rank optimization algorithm on top of the LRD's original rank selection method which searches in a domain around the original ranks and finds the candidates that lead to a more efficient calculation for each layer in the model. 

The proposed algorithm is explained as a pseudo code in Algorithm \ref{rank-selection}. 
As described here, the algorithm starts from the initial rank $R$ (which is calculated according to the desired compression ratio) and decreases it incrementally until it reaches to a the rank which leads to a lower computational time comparing to the original layer.
If it could not find such a rank with lower computational time, the original layer will be used instead. 
This is because for some layers, the original layer may be faster than the decomposed one. 


\begin{algorithm}
\caption{Find the rank $R_{opt}$ that leads to more efficient computations}
\label{rank-selection}
    \begin{algorithmic} 
    \STATE \textbf{Input}: Original layer $L$, Rank $R$, Lower bound rank $R_{\min}$, Input tensor $x$
    \STATE $T \leftarrow$ Processing time of original layer: $y = L(x)$
    \STATE \textbf{Initialization}: $r \leftarrow R$ and $R_{opt} \leftarrow 0$
    \WHILE{$r \geq R_{\min}$}
        \STATE $L_{r} \leftarrow$ Decompose layer $L$ using rank $r$
        \STATE $t(r) \leftarrow$ Processing time of decomposed layer: $y=L_{r}(x)$
            \IF{$r$ < $R$}
                \STATE $\Delta t(r) \leftarrow t(r)-t(r-1)$
            \ENDIF
        \STATE $r \leftarrow r-1$
    \ENDWHILE
    \STATE \textbf{Optimal Rank:} $R_{opt} \leftarrow \argmax\limits_{r \in [R_{\min}, R]} \Delta t(r)$
    \IF{$t(R_{opt})$ < $T$}
        \STATE Replace $L$ with $L_r$ 
    \ELSE
        \STATE Use original layer $L$
    \ENDIF
    \end{algorithmic}
\end{algorithm}

\begin{figure}
    \begin{minipage}{.48\textwidth}
        \centering
        \includegraphics[width=80mm,scale=1]{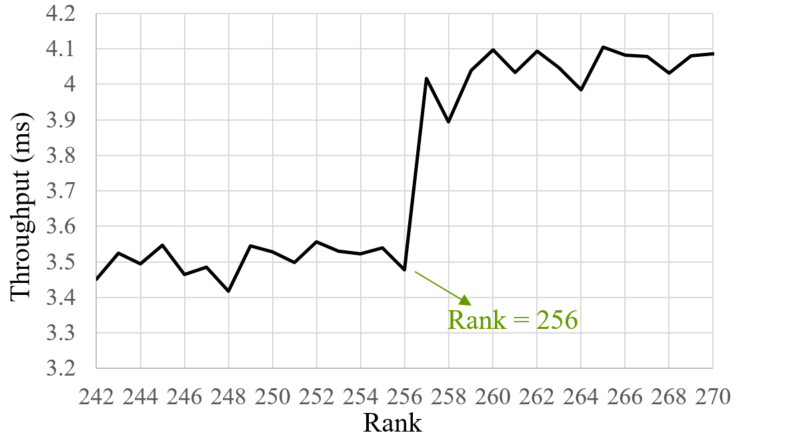}
        \subcaption{}
        \label{fig:2a}
    \end{minipage}
    \begin{minipage}{.48\textwidth}
        \centering
        \includegraphics[width=80mm,scale=1]{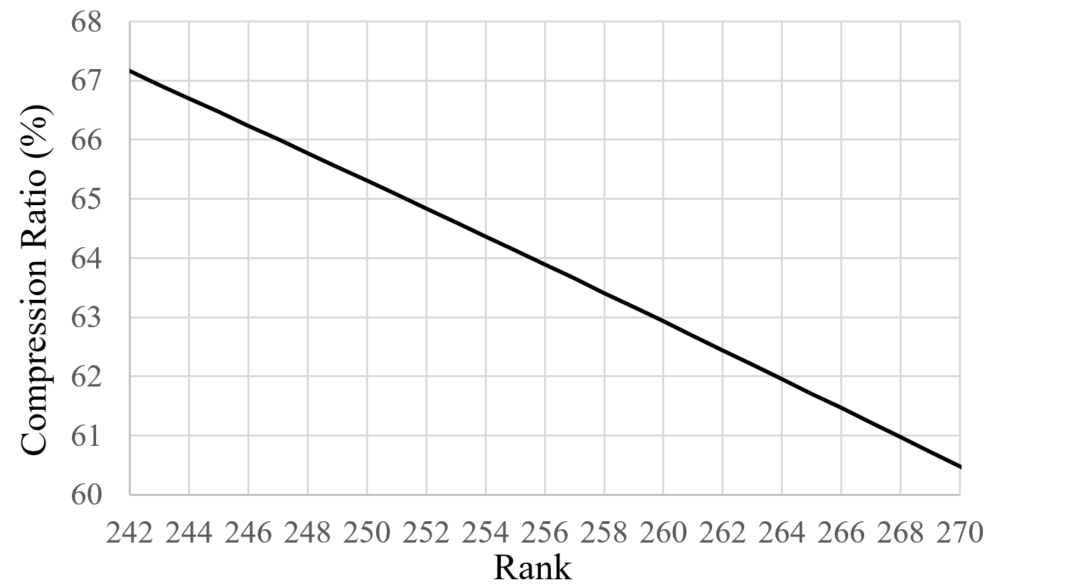}
        \subcaption{}
        \label{fig:2b}
    \end{minipage}
    \caption{Effect of rank selection on throughput of a 3x3 Conv layer in ResNet-152 with dimensions [512, 512, 3, 3] when decomposed using Tucker2 method with different ranks}
    \label{rank_selection}
\end{figure}

The ranks calculated by the proposed method are reported in Table \ref{resnet_ranks} for the beginning and late layers of ResNet-152 architecture. 
As seen in this table, some of the layers e.g. layer1.0.conv1 are not decomposed since the original layer is faster compared to the decomposed ones. 
The other layers are also decomposed using the ranks which result in a faster computation of the output. 
Note that here we used the PyTorch profiler for calculating the processing time of each layer. 
In Fig.\ref{rank_selection}, the effect of rank selection on the throughput of layers is depicted. 
As seen in this figure, changing the Tucker rank from 257 to 256 results in $15\%$ drop in throughput of the layer although the compression ratio stays almost the same (changes less than $1\%$). 

\begin{table}[h]
    \centering
    \caption{Ranks before and after rank optimization algorithm for early and late layers of ResNet-152 on Imagenet dataset }
{\footnotesize
    \begin{tabular}{l| c c c c}
{\bf Layer} &{\bf \# In Channels} &{\bf \# Out Channels} &{\bf 2x Ranks} &{\bf Optimized Ranks} \\
\hline
layer1.0.conv1 &64 &64 &16 &ORG \\
layer1.0.conv2 &64 &64 &38 &32 \\
layer1.0.conv3 &64 &256 &25 &24 \\
\hline
...\\
\hline
layer4.2.conv1 &2048 &512 &204 &202 \\
layer4.2.conv2 &512 &512 &309 &308 \\
layer4.2.conv3 &512 &2048 &204 &200 \\
fc &2048 &1001 &335 &253 \\
\end{tabular}
    }
    \label{resnet_ranks}
\end{table}

\subsection{Layer Freezing}
Another method we propose for accelerating the decomposed models is to fine-tune only one of the decomposed layers and freeze the rest of them. 
This is because the decomposed layers are calculated from the original layer using a low rank decomposition algorithm, assuming that decomposed weight tensors are close enough to the original weight tensors when they are reconstructed. 
Therefore, we can consider the frozen layers as transformation functions and hence, we may not need to update their weights during the optimization. 
To this end, we freeze the weights of the first 1x1 convolutional layer in Fig.\ref{fig:1a} and the first and last 1x1 convolutional layer in Fig.\ref{fig:1b}. 
This way we can save a lot of time during fine-tuning after decomposing the model.
However, note that although this method would accelerate the training phase, the inference speed would be the same as the vanilla LRD method as the number of layers and weights are the same during the inference phase. 

\subsection{Layer Merging}
The previous techniques proposed for accelerating the decomposed models still use the same number of layers as the vanilla LRD. 
In this section, we propose another approach which can result in a decomposed model with exactly the same number of layers as the original model but with much less number of parameters. 

\begin{figure}[h]
    \centering
    \includegraphics[width=\textwidth]{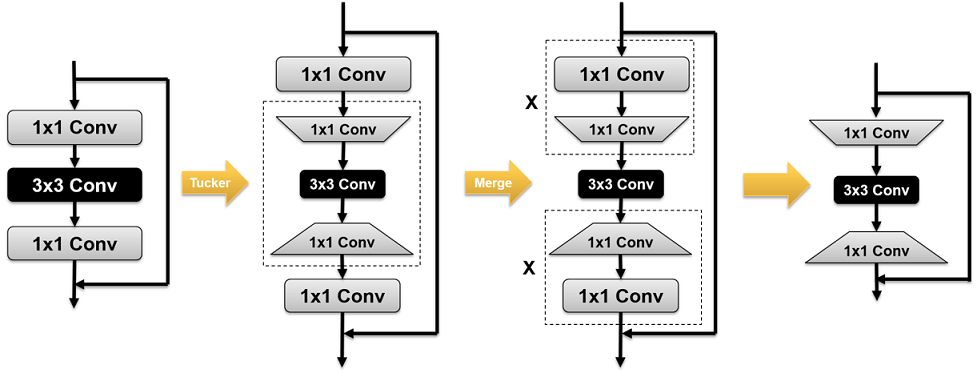}
    \caption{Mixing consecutive 1x1 Conv layers in ResNet modules after applying Tucker decomposition to the middle 3x3 Conv layer.}
    \label{fig:mix}
\end{figure}

\subsection{Branching Tucker}
In Fig.\ref{fig:1b}, it is shown how Tucker decomposition decomposes each kxk convolutional layer into 3 consecutive conv layers. 
As seen in this figure, Tucker uses 2 ranks for decomposition $r_1$ and $r_2$. 
These ranks are calculated so that a desired compression ratio is achieved.
For example, for a convolutional layer $\W\in \real^{C\times S\times h\times w}$, the ranks could be selected as follow to achieve a compression ratio of $\alpha$:
\begin{equation}\label{tucker_rank}
r_1=\frac{-\frac{C+\beta S}{\beta k^2} + \sqrt{\frac{(C+\beta S)^2}{\beta^2k^4} + \frac{4CS}{\beta \alpha}}}{2}
\end{equation}

However, there is always a trade-off between compression ratio and performance as using smaller ranks for more compression can result in significant drop in accuracy.
Therefore, we propose a more efficient way of implementing Tucker decomposition in multiple parallel branches so that with the same large ranks, we can reduce computational cost without compromising the accuracy.

Tucker decomposition shown in \eqref{tucker2} can be rewritten in two steps:
\begin{align}\label{tucker2-2}
\Y &= \X \times_C \U = \sum_{i=1}^{r_1}{\u_i\x_i^\top} \\
\W &= \Y \times_S \V = \sum_{i=1}^{r_2}{\y_i\v_i^\top}
\end{align}
in which $\Y\in \real^{C\times r_2\times k^2}$ is the result of multiplying the first 1x1 conv layer with the core 3x3 conv layer. 
However, assuming that the ranks $r_1$ and $r_2$ are quantized to multiples of integer $N$:
\begin{align}\label{split}
    r_1 = NR_1 \\
    r_2 = NR_2
\end{align}
According to \eqref{tucker2-2} we can write:

\begin{align}\label{tucker2-4}
\W &= \sum_{i=1}^{R_2}{\y_i\v_i^\top} + \sum_{i=R_2+1}^{2R_2}{\y_i\v_i^\top} + ... \sum_{i=(N-1)R_2+1}^{NR_2}{\y_i\v_i^\top} \\
&= \sum_{j=1}^{N}(\sum_{i=(j-1)R_2+1}^{jR_2}{\y_i\v_i^\top})\\
&= \sum_{j=1}^{N}(\Y_{j} \times_{R_2} \V_{j})
\end{align}
where $\Y_j\in \real^{C\times R_2\times k^2}$ and $\V_j\in \real^{R_2\times S}$ are truncated versions of $\Y$ and $\V$ which include the $j$th group of $R_2$ columns of $\Y$ and $\V$, respectively.
According to \eqref{tucker2-2}, assuming that $\X_j\in \real^{R_1\times R_2\times k^2}$ and $\U_j\in \real^{C\times R_1}$ are the truncated versions of $\X$ and $\U$ we have:
\begin{align}\label{tucker2-5-0}
\Y_j &= \sum_{i=(j-1)R_1+1}^{jR_1}{\u_i\x_i^\top} \\
&= \X_j\times_{R_1}\U_j
\end{align}
for $j=1,...,N$. 
Substituting \eqref{tucker2-5-0} into \eqref{tucker2-4} we have:
\begin{align}\label{tucker2-4}
\W &= \sum_{j=1}^{N}(\X_j\times_{R_1}\U_j \times_{R_2} \V_{j})
\end{align}



From this equation, it can be concluded that Tucker decomposition with ranks $r_1$ and $r_2$ can be split into $N$ parallel branches, each with smaller ranks of $R_1=r_1/N$ and $R_2=r_2/N$.
This way, we can reduce computational complexity without even reducing ranks for Tucker decomposition.  
We can also calculate the weights in each branch from the original weights.
This way we don't need to train from scratch. 

\begin{figure}[h]
    \centering
    \includegraphics[width=\textwidth]{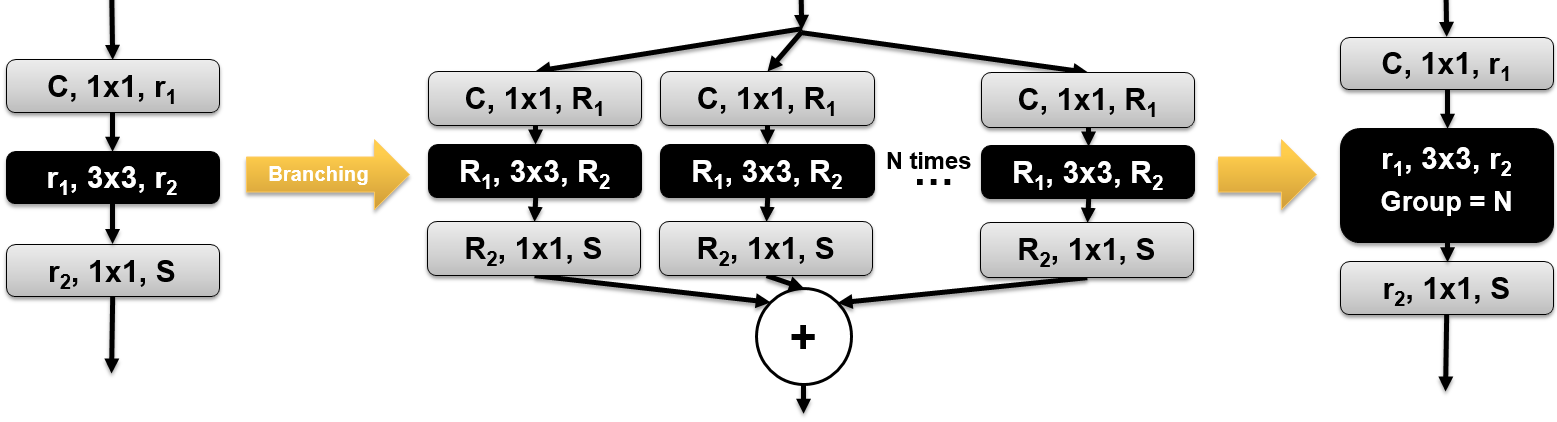}
    \caption{Branched Tucker decomposition and how it can be implemented efficiently using grouped convolutions.}
    \label{grouped}
\end{figure}

Fortunately, it can be shown that branched Tucker architecture can efficiently be implemented using grouped convolutions \cite{xie2017aggregated}.  
As depicted in Fig.\ref{grouped}, the last two architectures shown in this figure are equivalent.
According to this figure, the total number of parameters in the 3x3 conv layer inside the branched Tucker decomposition would be:
\begin{align}\label{tucker_num_params}
&= N\times (R_1\times R_2\times 9) \\
&= N\times (\frac{r_1}{N} \times \frac{r_2}{N}\times 9)\\
&= \frac{1}{N}\times (r_1 \times r_2\times 9)
\end{align}
Comparing to the the total number of parameters in the 3x3 conv layer of the vanilla Tucker i.e. $(r_1 \times r_2\times 9)$, it means that the layer could be compressed by $N$ times without even reducing the rank.

\begin{figure}[h]
    \centering
    \includegraphics[width=0.75\textwidth]{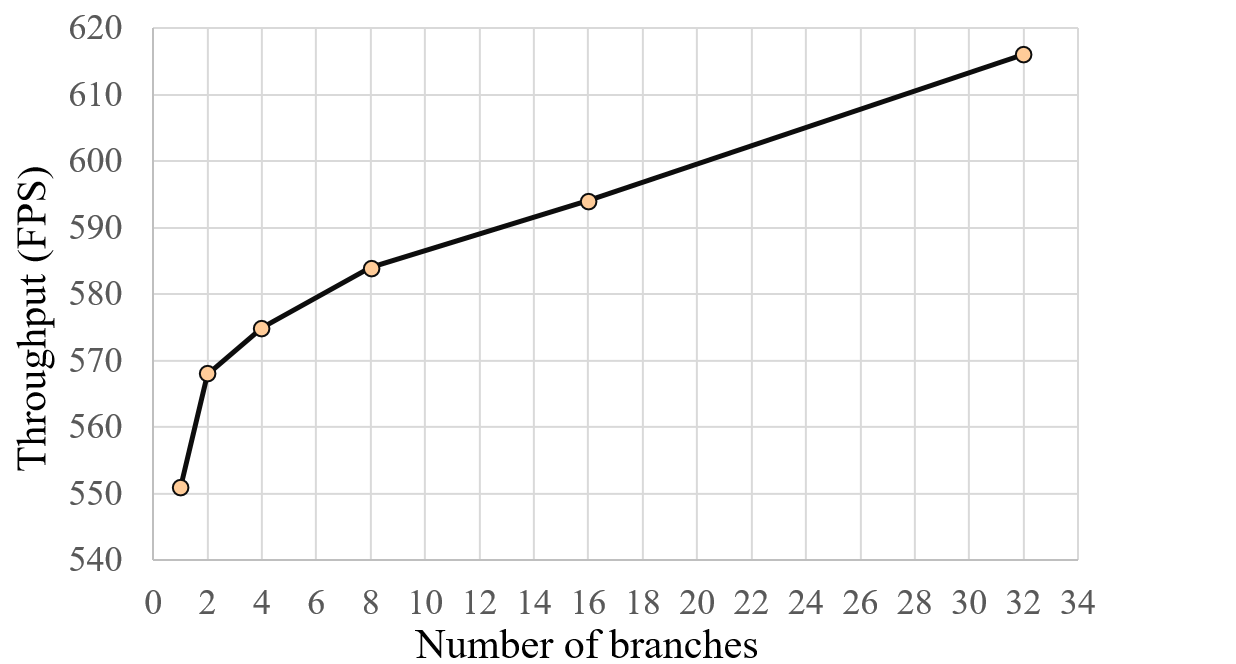}
    \caption{Throughput of the model vs number of branches in each layer for ResNet-152.}
    \label{grouped3}
\end{figure}

\section{Experimental Results}

\begin{table}[h]
    \centering
    \caption{Statistics of ResNet-50, ResNet-101 and ResNet-152 architectures before and after applying LRD.}
{\footnotesize
    \begin{tabular}{l| c c c c c}
{\bf Model} &{\bf Layers} &{\bf Comp Ratio} &{\bf $\Delta$FLOPs} &{\bf Train Speed-up} &{\bf Infer Speed-up}\\
\hline
\textbf{ResNet-50} & & & & &\\
Vanilla LRD &115 &-50.00 &-43.26 &+6.07 &+6.82\\
Optimized Ranks &115 &-47.28 &-47.71 &+14.16 &+13.96\\
Layer Freezing &115 &-50.00 &-43.26 &+24.57 &+6.82\\
Layer Merging &50 &-51.49 &-55.09 &+43.64 &+40.58\\
Layer Branching &0 &0 &0 &0 &0\\
\hline
\textbf{ResNet-101} & & & & &\\
Vanilla LRD &233 &-50.00 &-46.53 &+9.66 &+10.52\\
Optimized Ranks &233 &-51.96 &-49.56 &+14.49 &+19.92\\
Layer Freezing &233 &-50.00 &-46.53 &+29.95 &+10.52\\
Layer Merging &101 &-56.40 &-58.86 &+52.66 &+54.56\\
Layer Branching &0 &0 &0 &0 &0\\
\hline
\textbf{ResNet-152} & & & & &\\
Vanilla LRD &352 &-50.00 &-47.69 &+11.73 &+13.14 \\
Optimized Ranks &352 &-51.86 &-50.20 &+15.86 &+18.43\\
Layer Freezing &352 &-50.00 &-47.69 &+31.72 &+13.14\\
Layer Merging &152 &-58.11 &-60.18 &+55.86 &+54.90\\
Layer Branching &0 &0 &0 &0 &0
\end{tabular}
    }
    \label{table1}
\end{table}

\begin{table}[h]
    \centering
    \caption{Comparison of accuracy and efficiency for ResNet-50.}
{\footnotesize
    \begin{tabular}{l| c c| c c| c c}
{\bf Method} &{\bf Top-1} &{\bf $\Delta$Top-1} &{\bf Top-5} &{\bf $\Delta$Top-5} &{\bf $\Delta$FLOPs} &{\bf $\Delta$Throughput}\\
\hline
DCP\cite{} &74.95 &-1.06 &92.32 &-0.61 &55.6 &-\\
CCP\cite{} &75.21 &-0.94 &92.42 &-0.45 &54.1 &-\\
MetaPruning\cite{} &75.40 &-1.20 &- &- &51.2 &-\\
GBN\cite{} &75.18 &-0.67 &92.41 &-0.26 &55.1 &-\\
HRank\cite{} &74.98 &-1.17 &92.33 &-0.54 &43.8 &-\\
Hinge\cite{} &74.70 &-1.40 &- &- &-54.4 &-\\
DSA\cite{} &74.69 &-1.33 &92.45 &-0.80 &-50.0 &-\\
SCP\cite{} &75.27 &-0.62 &92.30 &-0.68 &-54.3 &-\\
LeGR\cite{} &75.70 &-0.40 &92.70 &-0.20 &-42.0 &-\\
NPPM\cite{} &75.96 &-0.19 &92.75 &-0.12 &-56.0 &-\\
\hline
Vanilla LRD &76.67 &+0.54 &- &- &-43.26 &+6.82\\
Optimized Ranks &- &- &- &- &-47.71 &+13.96\\
Layer Freezing &- &- &- &- &-43.26 &+6.82\\
Layer Merging &75.91 &-0.21 &92.91 &+0.04 &-55.09 &+40.58\\
Layer Branching &0 &0 &0 &0 &0 &0

\end{tabular}
    }
    \label{table2}
\end{table}

\begin{table}[h]
    \centering
    \caption{Comparison of accuracy and efficiency for ResNet-101.}
{\footnotesize
    \begin{tabular}{l| c c| c c| c c}
{\bf Method} &{\bf Top-1} &{\bf $\Delta$Top-1} &{\bf Top-5} &{\bf $\Delta$Top-5} &{\bf $\Delta$FLOPs} &{\bf $\Delta$Throughput}\\
\hline
Rethinking \cite{} &75.37 &-2.10 &- &- &-47.0 &-\\
IE \cite{} &77.35 &-0.02 &- &- &-39.8 &-\\
FPGM \cite{} &77.32 &-0.05 &93.56 &0.00 &-41.1 &-\\
NPPM\cite{} &77.83 &+0.46 &93.77 &+0.21 &-56.0 &-\\
\hline
Vanilla LRD &76.94 &-0.43 &93.40 &-0.14 &-46.53 &+13.14\\
Optimized Ranks &- &- &- &- &-49.56 &+18.43\\
Layer Freezing &- &- &- &- &-46.53 &+13.14\\
Layer Merging &76.55 &-0.82 &93.4 &-0.14 &-58.86 &+54.90\\
Layer Branching &76.67 &-0.70 &93.36 &-0.19 &0 &+7.43

\end{tabular}
    }
    \label{table3}
\end{table}

\begin{table}[h]
    \centering
    \caption{Comparison of accuracy and efficiency for ResNet-152.}
{\footnotesize
    \begin{tabular}{l| c c| c c| c c}
{\bf Method} &{\bf Top-1} &{\bf $\Delta$Top-1} &{\bf Top-5} &{\bf $\Delta$Top-5} &{\bf $\Delta$FLOPs} &{\bf $\Delta$Throughput}\\
\hline
Vanilla LRD &- &- &- &- &-47.69 &+13.14\\
Optimized Ranks &- &- &- &- &-50.20 &+18.43\\
Layer Freezing &77.83 &-0.48 &93.93 &-0.11 &-47.69 &+13.14\\
Layer Merging &77.86 &-0.44 &94.116 &+0.07 &-60.18 &+54.90\\
Layer Branching &77.97 &-0.34 &93.93 &-0.11 &-66.75 &0

\end{tabular}
    }
    \label{table4}
\end{table}

\section{Conclusion}
In this work, a progressive low rank decomposition method was used for compression of large transformer based language models. In contrast to many of state-of-the-art compression methods where intensive pre-training of the compressed model is necessary, progressive LRD can provide promising performance by compressing the model in the fine-tuning stage. This leads to reduction in the computation resources needed for obtaining a compressed model for a given task.
We show that in later steps of the iterative compression where the student models becomes much smaller than the teacher (compression factor larger than 8$\times$) KD can be used to improve the performance.


\bibliography{natbib}
\bibliographystyle{acl_natbib}

\end{document}


\appendix

\section{Appendix}

\subsection{Published and reproduced models}
We reproduce the state-of-the-art models for SF and ID. The resulting trained models obtain similar results to the published, as shown in Appendix Table \ref{tab:published_scores}.

\begin{table}[th]
    \begin{savenotes}
    \renewcommand{\arraystretch}{0.9}
    \setlength{\tabcolsep}{3pt}
        \centering
        \begin{tabular}{l|ll|ll|ll}
        \toprule
        \multirow{2}{*}{Test Set}  & \multicolumn{2}{c}{\bf ATIS} &\multicolumn{2}{c}{\bf SNIPS} &\multicolumn{2}{c}{\bf NLU-ED} \\
        & Slot & Int.& Slot & Int. \\
        \midrule
        \multicolumn{7}{c}{Stack-Prop+BERT} \\
        \midrule
        Published & 96.1 & 97.5 & 97.0 & 99.0 & na & na \\
        Reproduced & 95.7 & 96.5 & 95.0 & 98.2 & 74.0 & 85.1 \\
        \midrule
        \multicolumn{7}{c}{Bi-RNN} \\
        \midrule
        Published & 94.9 & 97.6 & 89.4* & 97.1* & na & na  \\
        Reproduced & 95.7 & 96.5 & 95.0 & 98.3 & 65.8 & 78.8  \\
        \bottomrule
        \end{tabular}
        \caption{Published and reproduced SF and ID results. The numbers with * indicate that the scores were not published in the original \cite{wang2018bi} paper but in \cite{qin2019stack}.}
        \label{tab:published_scores}
    \end{savenotes}
\end{table}

\subsection{Survey}
\label{sec:survey}
In Appendix Tables \ref{fig:survey_instructions} and \ref{fig:survey_excerpt} We show the instructions and an excerpt of the sentences, as presented to the surveyed participants\footnote{We asked the participants to rate the fluency of each utterance (from 1 to 5) in order to average it over the control utterances. Allowing us to establish the annotator capacity of our volunteer participants. We expected this metric to reflect the high quality of the cherry-picked control utterances. As expected, our participants score remained between 4.2 and 5 out of 5.}.\\

\begin{figure}[th]
\centering
\begin{subfigure}{0.44\textwidth}
  \centering
  \includegraphics[width=1.0\linewidth]{images/survey0.PNG}
  \caption{Print-screen of the survey instructions.}
  \label{fig:survey_instructions}
\end{subfigure}%
\begin{subfigure}{0.56\textwidth}
\begin{subfigure}{0.50\textwidth}
  \centering
  \includegraphics[width=0.99\linewidth]{images/survey2.PNG}
\end{subfigure}%
\begin{subfigure}{0.50\textwidth}
  \centering
  \includegraphics[width=1.0\linewidth]{images/survey3.PNG}
\end{subfigure}
\caption{Print-screen excerpts of the survey.}
\label{fig:survey_excerpt}
\end{subfigure}
\end{figure}

\begin{table}[th]
    \begin{savenotes}
    \renewcommand{\arraystretch}{0.9}
    \setlength{\tabcolsep}{3pt}
        \centering
        \begin{tabular}{p{2cm}|llllllllllllll}
        \toprule
        & \multicolumn{7}{c|}{Group 1} & \multicolumn{7}{c}{Group 2} \\
        Participant Id & 1 & 2 & 3 & 4 & 5 & 6 & 7 & 8 & 9 & 10 & 11 & 12 & 13 & 14 \\
        \midrule
        \multicolumn{15}{c}{Experiment} \\
        \midrule
        Slot & 95.3 & 96.9 & 95.3 & 91.3 & 94.5 & 96.1 & 92.1 & 86.1 & 98.3 & 98.2 & 95.7 & 90.4 & 97.4 & 90.4 \\
        Intent & 83.3 & 93.3 & 87.9 & 83.3 & 90.0 & 91.7 & 93.3 & 76.7 & 90.0 & 88.1 & 93.2 & 87.7 & 84.5 & 81.4 \\
        \midrule
        \multicolumn{15}{c}{Control} \\
        \midrule
        Fluency & 4.9 & 5 & 4.8 & 4.6 & 4.9 & 4.7 & 4.5 & 4.2 & 4.3 & 5 & 5 & 4.9 & 4.8 & 4.2 \\
        Slot & 89.5 & 89.5 & 100 & 94.7 & 100 & 94.7 & 89.5 & 94.7 & 100 & 100 & 100 & 100 & 94.7 & 89.5 \\
        Intent & 91.7 & 100 & 100 & 100 & 100 & 100 & 91.7 & 91.7 & 100 & 100 & 100 & 90.9 & 100 & 100 \\
        \bottomrule
        \end{tabular}
        \caption{Survey results and statistics per participant. The average slot score and the average intent score appear as percentages, the average sentence fluency score appears as a scale from 1 to 5.}
        \label{tab:survey_result_complete}
    \end{savenotes}
\end{table}

\subsection{Complete table of NATURE operators applied to ATIS, SNIPS and NLU-ED}
In the Appendix Tables \ref{tab:res_complete} and \ref{tab:bert_res_complete} we present all obtained scores ran on 2 models trained on the original train and validation sets of ATIS, SNIPS and NLU-ED and evaluated on the original, random and hard altered test sets.

\begin{savenotes}
\begin{table*}[th]
\renewcommand{\arraystretch}{0.9}
\setlength{\tabcolsep}{3pt}
    \small
    \centering
    \begin{tabular}{l|lll|lll|lll}
    \toprule
  \multirow{2}{*}{Test Set}  & \multicolumn{3}{c}{\textbf{ATIS}} & \multicolumn{3}{c}{\textbf{SNIPS}} & \multicolumn{3}{c}{\textbf{NLU-ED}} \\
  & \vtop{\hbox{\strut Slot}\hbox{\strut (F1)}} & \vtop{\hbox{\strut Intent}\hbox{\strut (Acc)}} & \vtop{\hbox{\strut E2E}\hbox{\strut (Acc)}} & \vtop{\hbox{\strut Slot}\hbox{\strut (F1)}} & \vtop{\hbox{\strut Intent}\hbox{\strut (Acc)}} &  \vtop{\hbox{\strut E2E}\hbox{\strut (Acc)}} & \vtop{\hbox{\strut Slot}\hbox{\strut (F1)}} & \vtop{\hbox{\strut Intent}\hbox{\strut (Acc)}} &  \vtop{\hbox{\strut E2E}\hbox{\strut (Acc)}} \\
  \midrule
  \multicolumn{10}{c}{Stack-Prop+BERT} \\
  \midrule
  Original & 95.7 & 96.5 & 86.2 & 95.0 & 98.3 & 87.9 & 74.0 & 85.1 & 67.8 \\
  Random & 91.3 & 95.0 & 66.5 & 83.4 & 96.1 & 53.8 & 67.4 & 76.1 & 56.8 \\
  & \quad $\pm$ 0.1 & \quad $\pm$ 0.3 & \quad $\pm$ 1.0 & \quad $\pm$ 0.5 & \quad $\pm$ 0.3 & \quad $\pm$ 3.2 & \quad $\pm$ 0.1 & \quad $\pm$ 0.2 & \quad $\pm$ 0.2 \\
  Hard & 82.3 & 90.7 & 34.9 & 70.6 & 95.3 & 12.9 & 55.5 & 62.7 & 38.9 \\
  \midrule
  \multicolumn{10}{c}{Bi-RNN} \\
  \midrule
  Original & 94.7 & 97.6 & 84.3 & 88.9 & 97.6 & 77.3 & 65.9 & 82.1 & 61.9 \\
  Random & 89.9 & 94.3 & 61.8 & 75.6 & 94.1 & 39.0 & 60.6 & 70.8 & 50.1 \\
  & \quad $\pm$ 0.1 & \quad $\pm$ 0.1 & \quad $\pm$ 1.6 & \quad $\pm$ 0.5 & \quad $\pm$ 0.1 & \quad $\pm$ 2.5 & \quad $\pm$ 0.4 & \quad $\pm$ 0.4 & \quad $\pm$ 0.3 \\
  Hard & 79.9 & 92.0 & 27.6 & 62.4 & 92.9 & 7.0 & 49.6 & 58.8 & 34.5 \\
  \bottomrule
\end{tabular}
\caption{Stack-Prop+BERT and Bi-RNN performances for ATIS, SNIPS and NLU-ED. We report F1 slot filling, accuracy for intent detection and end-to-end accuracy overall. The reported scores of the Random altered test set are a mean of 10 random distribution of processes and is accompanied by the variance score.}
\label{tab:res_complete}
\end{table*}
\end{savenotes}

\subsection{Complete NATURE operators applied to Data Augmented versions of ATIS, SNIPS and NLU-ED}
In the Appendix Table \ref{fig:nat_vs_da} we compare our NATURE operators and common automatic DA strategies from the NLPaug library. In the Appendix Table \ref{tab:da_res_complete} we present all obtained scores ran on 2 models trained on a Data Augmented version of ATIS, SNIPS and NLU-ED.

\begin{savenotes}
  \renewcommand{\arraystretch}{1.4}
  \setlength{\tabcolsep}{3pt}
    \centering
    \begin{table}
        \begin{tabular}{p{0.07\textwidth}|p{0.45\textwidth}|p{0.05\textwidth}|p{0.43\textwidth}}
        \toprule
        \multicolumn{4}{c}{\textbf{Original:} find a tv series called armageddon summer} \\
        \midrule
        \multicolumn{2}{c|}{\textbf{NATURE}} & \multicolumn{2}{c}{\textbf{DA}} \\
        \midrule
        BOS Filler & \textbf{yeah so} find a tv series called armageddon summer & Keyb. & find a tv \textbf{seriSs} called \textbf{armaRdvdon} summer \\
        PreV Filler & \textbf{basically} find a tv series called armageddon summer & Spell. & \textbf{fine} a tv \textbf{serie} called armageddon summer \\
        PosV Filler & find \textbf{you know} a tv series called armageddon summer & Syn. & find a tv \textbf{set} series called armageddon summertime \\
        EOS Filler & find a tv series called armageddon summer \textbf{if it pleases mi liege} & Ant. & \textbf{lose} a tv series called armageddon summer \\
        Syn. V. & \textbf{finds} a tv series called armageddon summer & TF IDF & find tv series called armageddon \textbf{forms} \\
        Syn. Adj. & find a tv series called \textbf{last} summer & Ctxt. WE. & find a \textbf{second} series called armageddon \textbf{ii} \\
        Syn. Adv. & find a \textbf{another} series called armageddon summer & & \\
        Syn. SW & find \textbf{and} tv series called armageddon summer & & \\
        Speako & find a tv \textbf{serie} called armageddon summer & & \\
        \bottomrule
        \end{tabular}
    \caption{Nature and DA candidates for the same utterance.}
    \label{fig:nat_vs_da}
    \end{table}
\end{savenotes}

\begin{savenotes}
\begin{table*}[th]
\renewcommand{\arraystretch}{0.9}
\setlength{\tabcolsep}{3pt}
    \small
    \centering
    \begin{tabular}{l|lll|lll|lll}
    \toprule
  \multirow{2}{*}{Test Set}  & \multicolumn{3}{c}{\textbf{ATIS}} & \multicolumn{3}{c}{\textbf{SNIPS}} & \multicolumn{3}{c}{\textbf{NLU-ED}} \\
  & \vtop{\hbox{\strut Slot}\hbox{\strut (F1)}} & \vtop{\hbox{\strut Intent}\hbox{\strut (Acc)}} & \vtop{\hbox{\strut E2E}\hbox{\strut (Acc)}} & \vtop{\hbox{\strut Slot}\hbox{\strut (F1)}} & \vtop{\hbox{\strut Intent}\hbox{\strut (Acc)}} &  \vtop{\hbox{\strut E2E}\hbox{\strut (Acc)}} & \vtop{\hbox{\strut Slot}\hbox{\strut (F1)}} & \vtop{\hbox{\strut Intent}\hbox{\strut (Acc)}} &  \vtop{\hbox{\strut E2E}\hbox{\strut (Acc)}} \\
  \midrule
  \multicolumn{10}{c}{Stack-Prop+BERT} \\
  \midrule
  Original & 94.7 & 95.7 & 83.3 & 93.8 & 97.7 & 85.3 & 72.4 & 83.8 & 66.2 \\
  Random & 91.7 & 94.3 & 69.2 & 85.7 & 96.0 & 64.4 & 67.3 & 75.6 & 56.7 \\
   & \quad $\pm$ 0.0 & \quad $\pm$ 0.1 & \quad $\pm$ 0.9 & \quad $\pm$ 0.2 & \quad $\pm$ 0.4 & \quad $\pm$ 1.5 & \quad $\pm$ 0.2 & \quad $\pm$ 0.1 & \quad $\pm$ 0.2 \\
  Hard & 87.2 & 91.0 & 54.0 & 72.7 & 95.1 & 27.1 & 55.3 & 64.0 & 40.7 \\
  \midrule
  \multicolumn{10}{c}{Bi-RNN} \\
  \midrule
  Original & 93.7 & 96.9 & 81.8 & 86.2 & 97.6 & 69.7 & 66.3 & 82.5 & 61.8 \\
  Random & 90.3 & 93.9 & 65.6 & 77.4 & 95.3 & 48.2 & 61.2 & 73.4 & 51.8 \\
  Random & \quad $\pm$ 0.1 & \quad $\pm$ 0.2 & \quad $\pm$ 1.1 & \quad $\pm$ 0.3 & \quad $\pm$ 0.2 & \quad $\pm$ 1.8 & \quad $\pm$ 0.1 & \quad $\pm$ 0.2 & \quad $\pm$ 0.2 \\
  Hard & 83.2 & 92.8 & 43.0 & 65.0 & 94.1 & 19.1 & 62.1 & 50.2 & 38.6 \\
  \bottomrule
\end{tabular}
\caption{Stack-Prop+BERT and Bi-RNN performances for ATIS, SNIPS and NLU-ED using data augmentation on the train and validation sets. We report F1 slot filling, accuracy for intent detection and end-to-end accuracy overall. The reported scores of the Random altered test set are a mean of 10 random distribution of processes and is accompanied by the variance score.}
\label{tab:da_res_complete}
\end{table*}
\end{savenotes}

\bibliography{natbib}
\bibliographystyle{acl_natbib}